\documentclass[%
 reprint,
 amsmath,amssymb,
 aps,
]{revtex4-2}

\usepackage[colorlinks,linktocpage=true]{hyperref}	
\usepackage{graphicx}% Include figure files
\usepackage{dcolumn}% Align table columns on decimal point
\usepackage{bm}% bold math
\renewcommand{\v}[1]{\ensuremath{\mathbf{#1}}} % for vectors
\newcommand{\gv}[1]{\ensuremath{\mbox{\boldmath$ #1 $}}} 
\newcommand{\abs}[1]{\left| #1 \right|} % for absolute value
\renewcommand\eqref[1]{Eq.\;\ref{#1}} % new version of eqref
\hypersetup{						
  citecolor= black,
  linkcolor=black,
  urlcolor=black,
}

\usepackage [english]{babel}
\usepackage [english = american]{csquotes}
\MakeOuterQuote{"}

\usepackage{color}					% Allows picking of custom RGB colors

\begin{document}

\title{Model scale versus domain knowledge in statistical forecasting of chaotic systems}

\author{William Gilpin}
 \email{wgilpin@utexas.edu}
\affiliation{%
Department of Physics, The University of Texas at Austin, Austin, Texas 78712, USA
}%
\affiliation{%
Oden Institute for Computational Engineering and Sciences, The University of Texas at Austin, Austin, Texas 78712, USA
}%

\date{\today}% It is always \today, today,

\newpage
\begin{abstract}
Chaos and unpredictability are traditionally synonymous, yet large-scale machine learning methods recently have demonstrated a surprising ability to forecast chaotic systems well beyond typical predictability horizons. However, recent works disagree on whether specialized methods grounded in dynamical systems theory, such as reservoir computers or neural ordinary differential equations, outperform general-purpose large-scale learning methods such as transformers or recurrent neural networks. These prior studies perform comparisons on few individually-chosen chaotic systems, thereby precluding robust quantification of how statistical modeling choices and dynamical invariants of different chaotic systems jointly determine empirical predictability. Here, we perform the largest to-date comparative study of forecasting methods on the classical problem of forecasting chaos: we benchmark $24$ state-of-the-art forecasting methods on a crowdsourced database of $135$ low-dimensional systems with $17$ forecast metrics. We find that large-scale, domain-agnostic forecasting methods consistently produce predictions that remain accurate up to two dozen Lyapunov times, thereby accessing a new long-horizon forecasting regime well beyond classical methods. We find that, in this regime, accuracy decorrelates with classical invariant measures of predictability like the Lyapunov exponent. However, in data-limited settings outside the long-horizon regime, we find that physics-based hybrid methods retain a comparative advantage due to their strong inductive biases.
\end{abstract}

\maketitle

\newpage
\section{Introduction}

Chaos traditionally implies the butterfly effect: a small change in a system grows exponentially over time, complicating efforts to reliably forecast the system's long-term evolution. Predicting chaos therefore represents a longstanding problem at the interface of physics and computer science \cite{boffetta2002predictability}, even motivating early applications of artificial neural networks during the 1991 Santa Fe forecasting competition \cite{weigend1993results}. Recent successes in statistical forecasting motivate revisiting this problem, by providing compelling examples of data-driven prediction of diverse systems such as cellular signaling pathways \cite{yazdani2020systems}, hourly precipitation forecasts \cite{espeholt2022deep}, active nematics \cite{colen2021machine}, and tokamak plasma disruptions \cite{zheng2018hybrid}. 

However, there is little consensus whether the practical success of emerging forecasting methods stems from fundamental advances in representing and parameterizing chaos, or simply from the availability of larger datasets, model capacities, and computational resources \cite{lim2021time,tang2020introduction,godahewa2021monash}. Recent fundamental advances in representing chaos include works demonstrating that chaotic systems appear more linear when lifted to higher-dimensional representations than strictly necessary to describe their dynamics---such as those implicitly learned by large, overparameterized learning methods \cite{brunton2017chaos,wang2023koopman,otto2021koopman}. These works partly explain the recent emergence of reservoir computers as strong forecasting methods for dynamical systems \cite{gauthier2021next,bompas2020accuracy,platt2021robust,pathak2018model,jiang2019model,chattopadhyay2020data,vlachas2020backpropagation}; these models use emergent properties of random networks to lift complex time series into a random feature space, thereby simplifying learning at the expense of increasing model complexity \cite{maass2002real,jaeger2004harnessing}. Other recently-introduced hybrid models directly encode dynamical constraints within their model formulation \cite{karniadakis2021physics}; among these are neural ordinary differential equations \cite{chen2018neural}, physics-informed neural networks \cite{raissi2019physics}, and recurrent neural networks with domain-specific architectural modifications \cite{sangiorgio2020robustness,bhat2022recurrent,yu2017learning}. Broadly, these physics-based models can be seen as containing inductive biases---architectural or modeling choices informed by knowledge that the target time series are drawn from dynamical systems---that effectively reduce the variance of potential fitted models in exchange for more efficient training \cite{gilpin2023generative}.

\begin{figure*}[ht]
{
\centering
\includegraphics[width=\linewidth]{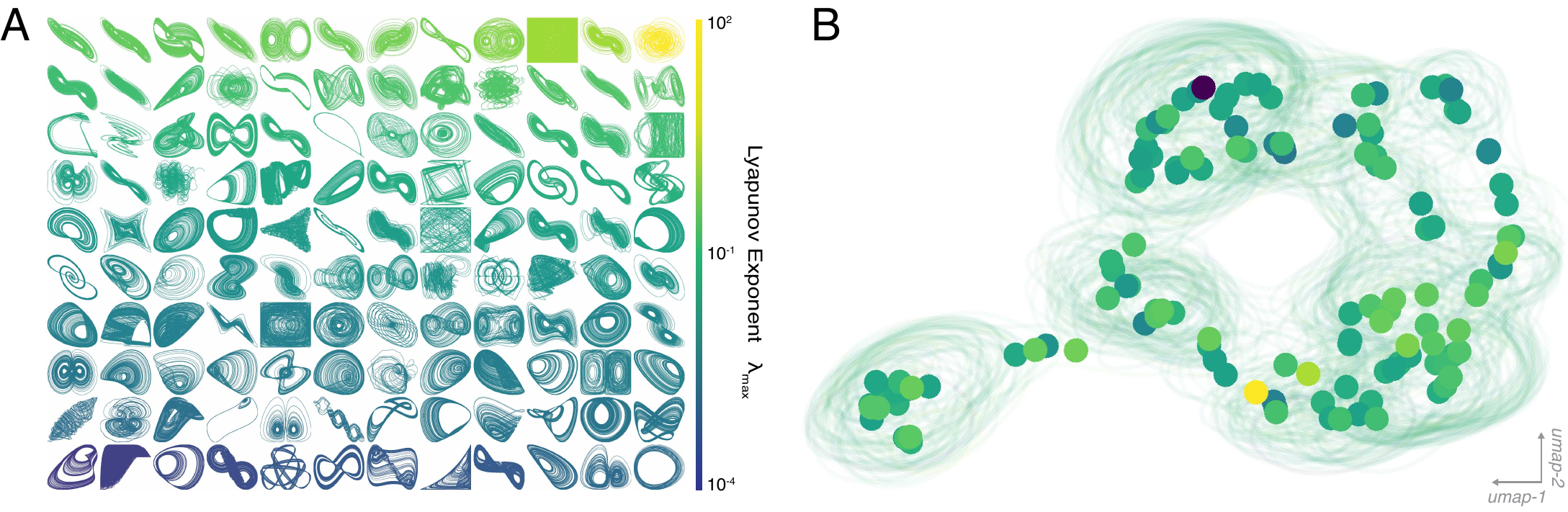}
\caption{
{\bf A space of low-dimensional chaotic systems.} (A) A dataset of $135$ distinct low-dimensional chaotic systems, colored by largest Lyapunov exponent ($\lambda_\text{max}$). (B) A nonlinear embedding of the attractors. Each attractor is featurized using $747$ invariant properties such as entropy, fractal dimension, et al., and then embedded in a two-dimensional vector space with UMAP. Contours denote $50\%$ confidence intervals in each system's embedding across $500$ random initial conditions and feature subsets; points denote centroids for each system.
}
\label{orbit}
}
\end{figure*}

In contrast, recent works by the computer science community advocate training large models with minimal domain-specific inductive biases \cite{elsayed2021we,lim2021time,tang2020introduction}. In controlled experiments, state-of-the art forecasting results have been achieved by large, overparametrized statistical learning models, such as transformers and hierarchical neural network architectures \cite{zhou2021informer,challu2023nhits,godahewa2021monash}. In principle, these models leverage their scale and the availability of large time series datasets to overcome lack of domain knowledge, and they have demonstrated consistently improving performance in major time series forecasting competitions and benchmarks \cite{hyndman2018forecasting,hewamalage2021look,makridakis2020m4}. When these domain-agnostic models have previously been applied to time series generated by chaotic dynamical systems, these models exhibit strong performance when sufficient training data is available \cite{chattopadhyay2020data,vlachas2020backpropagation}.

Complicating comparison of domain-agnostic versus physics-based chaotic forecasting models is a lack of systematic comparison on the same datasets. Prior works have compared methods on a handful of well-known chaotic attractors like the Mackey-Glass or Lorenz equations, or have used small domain-specific time series datasets like weather or biomedical time series \cite{makridakis2022m5,costa2019adaptive,godahewa2021monash,yu2017learning,bai2018empirical,yik2023neurobench}. A larger set of representative systems, and a controlled comparison among methods, is necessary to disentangle the relationship between chaoticity, predictability, and forecasting model architecture, as well as to understand how the properties of different black-box machine learning methods interact with the systems they predict.

Here, we systematically quantify the relationship between chaos and empirical predictability in a large-scale controlled experiment. We introduce a large-scale dataset of $135$ distinct low-dimensional chaotic attractors. For each system, we benchmark $24$ forecasting methods using $17$ forecast metrics that quantify both pointwise accuracy, and ability to capture invariant properties of the underlying attractors. When sufficient training data is available, we find that large, domain-agnostic forecasting models outperform physics-based models at both short and long forecasting horizons. However, when limitations are imposed on computational resources or data availability, models with inductive biases---particularly reservoir computers---perform more strongly. We find that invariant properties of the underlying dynamical systems only weakly correlate with the ability of the best-performing forecast models to forecast them, suggesting that scale and dataset availability, rather than intrinsic dynamical properties, limit the current ability of large models to forecast chaos.

\section{Methods}

\subsection{The chaotic attractors benchmark dataset}

We introduce a benchmark dataset containing $135$ low-dimensional differential equations describing known chaotic attractors \cite{gilpin2021chaos}. Originally curated from published works to include well-known systems such as the Lorenz, R\"ossler, and Chua attractors, since initial release the dataset has grown through crowdsourcing to include examples spanning diverse domains such as climatology, neuroscience, and astrophysics. Each dynamical system is aligned with respect to its dominant timescale and integration timestep using surrogate significance testing \cite{kantz2004nonlinear}, and is annotated with calculations of its invariant properties such as the Lyapunov exponent spectrum, fractal dimension, and metric entropy (\ref{a_dataset}).

Although each system has a distinct largest Lyapunov exponent ($\lambda_\text{max}$) indicating its putative chaoticity, some systems are closely related to each other. For example, the $19$ members of the Sprott attractor subfamily ($\lambda_\text{max} \in [0.01, 1.1]$) exhibit similar qualitative structure such as paired lobes, owing to the presence of predominantly quadratic nonlinearities in the governing differential equations \cite{sprott1994some,kaptanoglu2023benchmarking}. To identify relationships among attractors in our dataset, we convert each dynamical system into a high-dimensional vector by first generating a long trajectory, and then computing $747$ characteristic mathematical signal properties such as the metric entropy, power spectral coefficients, Hurst exponents, and others that are invariant to the initial conditions and sampling rate \cite{christ2018time,makowski2021neurokit2} (\ref{a_embedding}). We then use uniform manifold approximation and projection (UMAP) to visualize these high-dimensional vectors in a two-dimensional plane (Fig. \ref{orbit}) \cite{mcinnes2018umap,fulcher2013highly}. The resulting space of chaotic systems shows clear structure, with the Sprott and other scroll-like subfamilies clustering together, while qualitatively distinct systems separate. These results suggest chaoticity $\lambda_\text{max}$ represents just one among many invariant properties that relate different chaotic systems, as $\lambda_\text{max}$ correlates only weakly with the embedding ($\rho = 0.15 \pm 0.03$, bootstrapped Spearman rank-order coefficient). This visualization allows us to consider our dynamical systems dataset not as merely a list of differential equations, but rather as a set of points in a space of dynamical systems parametrized by their invariant properties 

\subsection{Forecasting models evaluated.}

We evaluate $24$ statistical forecasting models across all $135$ dynamical systems. We choose forecasting methods representing the broad diversity of methods available in the recent literature \cite{lim2021time,herzen2022darts}. Traditional methods include standard linear regression, autoregressive moving averages (ARIMA), exponential smoothing, Fourier mode extrapolation, boosted random forest models \cite{kantz2004nonlinear}, and newly-introduced linear models that account for trends and distribution shift \cite{zeng2022transformers}. Current state-of-the-art models for general time series forecasting are based on deep neural networks: the transformer model \cite{zhou2021informer}, long-short-term-memory networks (LSTM), vanilla recurrent neural networks (RNN), temporal convolutional neural networks \cite{bai2018empirical}, and neural basis expansion/neural hierarchical interpolation (NBEATS/NHiTS) \cite{oreshkin2020nbeats,challu2023nhits}. The latter methods generate forecasts hierarchically by aggregating separate forecasts at distinct timescales (NBEATS), and can explicitly coarse-grain the time series to further reduce computational costs (NHiTS). We also consider hybrid physics-motivated methods such as neural ordinary differential equations \cite{chen2018neural}, which approximate the continuous-time differential equation underlying time series; and echo-state networks (ESN), which train a linear model on a fixed "reservoir" of random nonlinearities \cite{maass2002real,jaeger2004harnessing}. We include nonlinear vector autoregressive models (nVAR), a generalization of classical ESN that removes the need for an explicit reservoir---hence their designation as \textit{next-generation reservoir computers} \cite{gauthier2021next}. In order to provide reference values for observed scalings, we also include several naive models that underfit the data, including naive mean and simple seasonal estimators, as well as a Kalman filter with internal state frozen at its value at the end of training data availability \cite{hyndman2018forecasting,durbin2012time}.

\subsection{Forecasting benchmark design}

Our dynamical systems dataset allows us to systematically compare the forecasting ability of different statistical forecasting methods across diverse dynamical systems. Forecasting dynamical systems from observations is a well-established field \cite{kantz2004nonlinear}, and we structure our experiments as a standard long-term autoregressive forecasting task \cite{vlachas2023learning}. For each $D$-dimensional dynamical system, we generate two time series arising from distinct initial conditions on the system's attractor $\v{y}_\text{train}(t'), \v{y}_\text{test}(t') \in \mathbb{R}^{D}$, $t' \in [0, T]$, which we subdivide into past and future series at a time $t^* \in (0, T)$. The model parameters are first fit using $\v{y}_\text{train}(t')$ on the interval $t' \in [0, t^*]$, and the accuracy of the resulting predictions $\hat{\v{y}}_\text{train}$ relative to the true values $\v{y}_\text{train}$ on the remaining interval $t' \in (t^*, T]$ are used for model selection and hyperparameter tuning. Next, the final model from each model class is fit to $\v{y}_\text{test}(t')$ on the interval $t' \in [0, t^*]$, and the resulting future predictions $\hat{\v{y}}_\text{test}(t')$ are compared against the as-yet unseen true values $\v{y}_\text{test}(t')$ on the interval $t' \in (t^*, T]$, producing an error score $\epsilon_{ik}(t)$ representing the performance of the $i^{th}$ forecasting model on the $k^{th}$ dynamical system at forecast horizon $t \in (0, T-t^*]$ after the end of training data availability (note that $t \equiv t' - t^*$). We compute $17$ different error metrics, including root mean-squared error, pointwise correlation, mutual information, and Granger causality. We report our results in the main text in terms of the symmetric mean absolute percent error (sMAPE) $\epsilon_{ik}(t) \equiv 2/(t - t^*) \int_{t^*}^t (| \v{y}_{\text{test},k}(t') - \hat{\v{y}}_{\text{test},ik}(t')|) / (|\v{y}_{\text{test},k}(t')| + |\hat{\v{y}}_{\text{test},ik}(t')|) \,dt'$ due to its common use, conceptual simplicity, and correlation with other metrics \cite{godahewa2021monash,makridakis2022m5,gilpin2023recurrences,gilpin2021chaos,wang2023koopman}.

Each forecasting method represents a class of possible models parameterized by choices made regarding architecture (model size, number of layers, units per layer, activation function) or training (optimization epochs, batch sizes). Such choices can strongly affect the performance of different methods \cite{platt2021robust,kantz2004nonlinear,godahewa2021monash}, yet different forecasting methods do not necessarily have equivalent adjustable hyperparameters. For all methods, we initialize all hyperparameters at their default values used in the publications from which they were drawn. We use either the original authors' code when available, or widely-used reference implementations based on the original works \cite{herzen2022darts}. However, because prior works primarily feature individual systems like the Lorenz attractor or Mackey-Glass equations, we perform additional hyperparameter tuning for each dynamical system and forecasting method pair. Because different methods have different hyperparameters, we restrict hyperparameter tuning to the equivalent of the lookback window $T_\ell$ for each model. $T_\ell$ corresponds to the input size for deep neural networks, the number of features for random forests, the lag order of autoregressive models, the inverse leakage rate for reservoir computers, or the time lag for state space models. Importantly, while $t^*$ determines the total history length available for training, $T_\ell < t^*$ affects both training and inference; it effectively determines how many past timepoints are simultaneously used as inputs, which the model processes to output a prediction. 

Several timescales characterize our benchmark design: the \textit{lookback window} $T_\ell$ is a tunable hyperparameter for each forecasting method that corresponds to the number of past timepoints seen simultaneously by a model at a given time; the \textit{history length} $t^* \geq T_\ell$ represents the total number of timepoints available to learn the model's parameters during training. $t^*$ is typically several times larger than $T_\ell$, because fitting model parameters requires supervised training on several subsets of length $T_\ell$ drawn from the available history. The \textit{forecast horizon} $t$ represents the number of unseen timepoints into the future that are predicted autoregressively; and the \textit{Lyapunov time} $\lambda_\text{max}^{-1}$ is an invariant property of each distinct dynamical system representing the characteristic timescale over which forecasts are expected to lose accuracy due to the butterfly effect. We report our forecast results scaled by this quantity, in units of $\lambda_\text{max} t$.

Our long-term forecasting experiments require $\sim\!10^{19}$ floating point operations for training, model selection, and hyperparameter tuning, a figure comparable to the scale of other recent large-scale machine learning benchmarks \cite{canziani2016analysis}. We have previously validated our experiment design in a smaller-scale trial univariate study \cite{gilpin2021chaos}; this new, larger-scale computational study applies the same experiment design to larger and more varied benchmark models, more dynamical systems, and an order-of-magnitude longer multivariate forecasts. For brevity, we highlight results for the best-performing forecasting models, and defer the full tabular results and alternative accuracy metrics to \ref{a_results} and our open-source repository. 

\section{Results}

\subsection{Large, domain-agnostic time series models effectively forecast diverse chaotic systems}

\begin{figure}[!]
{
\centering
\includegraphics[width=\linewidth]{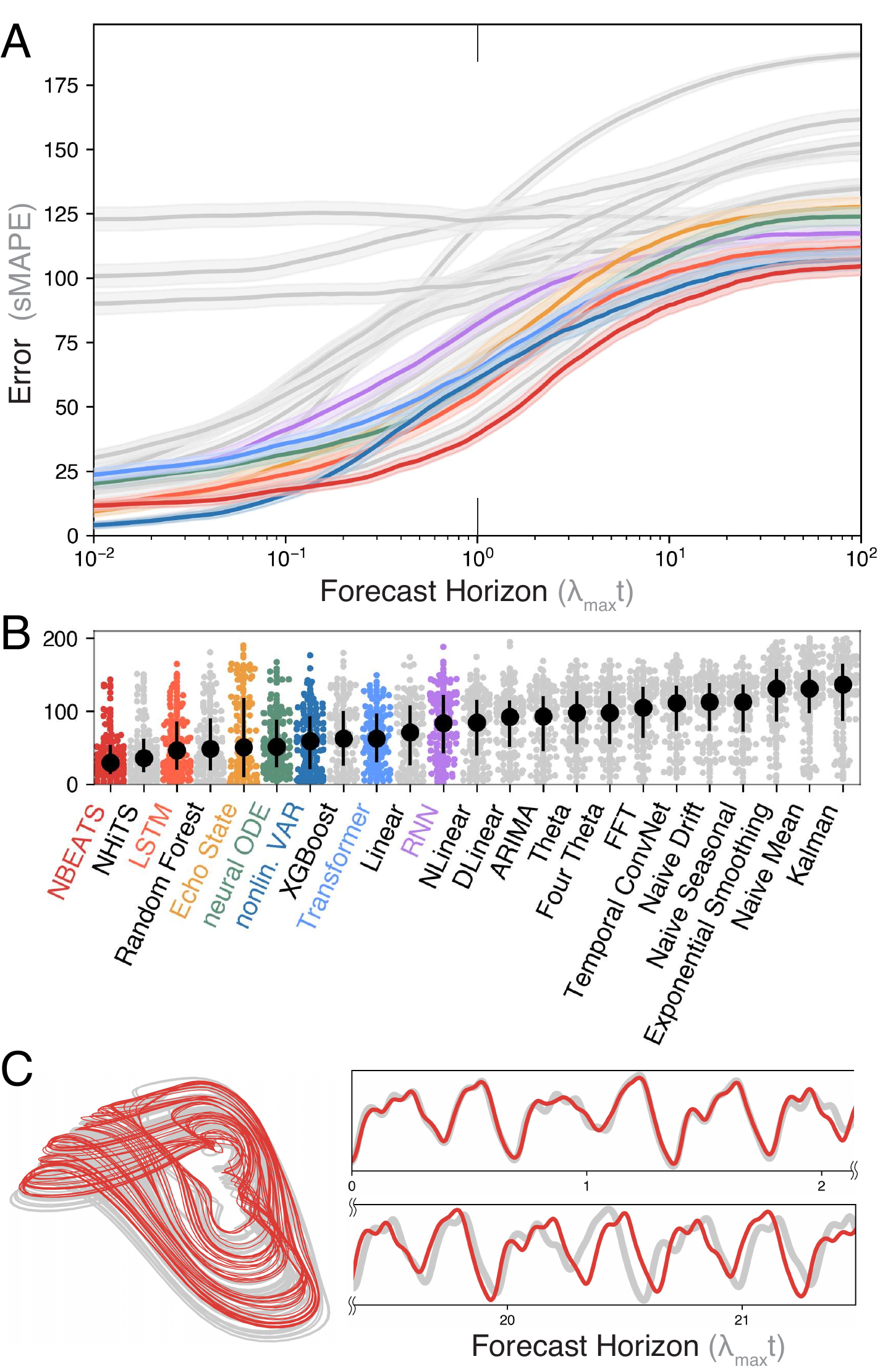}
\caption{
{\bf Statistical forecasting across an ensemble of chaotic systems.}  (A) The average error of $24$ forecasting methods $\langle \epsilon_{ik}(t) \rangle_k$ as a function of Lyapunov time, averaged across $135$ distinct chaotic systems. Colors denote high-performing models with properties of particular interest. (B) Distributions of the forecast errors when $t = \lambda_\text{max}^{-1}$. (C) The predictions of the best-performing forecast model (red), relative to a held-out true trajectory from the Mackey-Glass model (gray) at short and long forecasting horizons.
}
\label{forecast}
}
\end{figure}

Our main results are summarized in Figure \ref{forecast}. We observe that modern statistical learning methods successfully forecast diverse chaotic systems, with the strongest methods consistently succeeding across diverse systems and forecasting horizons. We highlight the strong relative performance of machine learning models in Figure \ref{forecast}C, where NBEATS successfully forecasts the Mackey-Glass equation for $\sim\!\! 22$ Lyapunov times without losing track of the global phase (see Fig. S1 for additional examples). Across all systems, the best-performing models achieve an average prediction time equal to $14 \pm 2 \,\lambda_\text{max}^{-1}$. These results extend beyond the $\sim\!10\,\lambda_\text{max}^{-1}$ reported for single systems in recent works \cite{pathak2018model,vlachas2023learning}, and sharply improves on the $\sim\!5\,\lambda_\text{max}^{-1}$ typically achieved before the widespread adoption of large machine learning models \cite{kantz2004nonlinear}. Our results underscore rapid progress since the $\sim\!1\,\lambda_\text{max}^{-1}$ targeted in the original Santa Fe competition \cite{weigend1993results}. The strongest-performing methods include NBEATS, NHiTS, transformers, and LSTM, which are all large models originally designed for generic sequential datasets, and which do not assume that input time series arise from a dynamical system. This observation suggests that more flexible, generic architectures may prove preferable for problems where some physical structure is present (e.g., analytic generating functions in the form of ordinary differential equations) but stronger domain knowledge (e.g., symmetries or symplecticity) is unavailable to further constrain learning. However, we note that the vanilla RNN and Temporal Convolutional Neural network exhibit unexpectedly weak performance, particularly given the latter model's relationship to NHiTS and the strong performance of the LSTM \cite{lea2017temporal,challu2023nhits,bai2018empirical}. Inspection of individual forecasts shows that both models exhibit instability at long forecast horizons, in which they quickly diverge from the underlying attractor and rapidly accrue errors. The chaotic nature of the time series amplifies this weakness compared to traditional time series forecasting benchmarks.

Among the remaining forecasting models, we note that the naive baselines perform poorly, as expected. Because chaotic systems are ergodic and thus statistically stationary over long intervals, baseline models that include components for constant linear drift perform particularly poorly: these models tend to fit a small but nonzero constant drift term given finite training data, which later causes the models to linearly diverge from the bounded attractor set during testing \cite{hyndman2018forecasting}. This effect explains the uncharacteristically low performance of the frozen Kalman model and exponential smoothing, two model that often perform well in short-horizon forecasting tasks \cite{makridakis2020m4,durbin2012time,hewamalage2021look,hyndman2018forecasting} where extrapolating the most recent monotonic trend plays a more significant role in determining aggregate accuracy. Conversely, among the classical models, the Fourier regression, ARIMA, and linear models perform comparatively well, due to their ability to model oscillating time series. These results underscore the unique aspects of our long-horizon chaos forecasting task, where models must correctly anticipate turning points and non-monotonic changes due to underlying attractor geometry. 

The strong performance of NBEATS/NHiTS suggests that this model has structural features favoring the chaotic systems dataset. Prior work has shown that hierarchical forecasting methods can flexibly integrate information across multiple timescales in a manner inaccessible to classical statistical models \cite{challu2023nhits}. While chaotic systems exhibit continuous spectra and thus contain information relevant to forecasting at a variety of timescales, many systems exhibit topologically-preferred timescales such as unstable periodic orbits---like the "loops" on either side of the Lorenz attractor---that dominate the system's underlying measure \cite{cvitanovic1988invariant}, and which therefore may represent higher priority motifs for learning. Highly performant model architectures therefore likely contain implicit inductive biases that advantage them on chaotic systems relative to other time series. Consistent with this finding, we note that reservoir computers (nVAR/ESN) also perform strongly on the chaotic systems dataset \cite{maass2002real,jaeger2004harnessing}, in agreement with prior observations for individual chaotic systems \cite{gauthier2021next,bompas2020accuracy,platt2021robust,pathak2018model,jiang2019model,chattopadhyay2020data,vlachas2020backpropagation}.

We contrast our results with recent reservoir computing studies that consider a subset of our $24$ forecast methods and $135$ chaotic systems \cite{shahi2022prediction,han2019review,vlachas2020backpropagation}. The comparatively strong performance of nVAR on our dataset likely stems from the quadratic nonlinearities used within the default set of fixed nonlinear kernels used by this method. Because most chaotic systems in our dataset feature predominantly quadratic nonlinearities, recent works show that the fully-trained nVAR can effectively learn to implement an exact multistep integrator for the dynamics \cite{zhang2023catch}. Moreover, while some recent works suggest that ESN/nVAR systematically outperform large-scale models on chaotic time series, we show below that larger domain-agnostic models regain their advantage when given sufficient training data. In contrast to prior works, we emphasize that our study performs model selection with respect to an equivalent hyperparameter found in all forecast methods, the lookback window $T_\ell$. When this parameter is left untuned at $T_\ell = 1$, machine learning methods like RNN or Transformers cannot use past context to inform their predictions, a condition analogous to creating a memoryless ESN/nVAR model by setting the leakage rate equal to one. This distinction potentially explains the weak relative performance of deep learning models compared to reservoir-based methods in recent chaotic forecasting benchmarks \cite{vlachas2020backpropagation,shahi2022prediction}. Additionally, while some ESN variants essentially represent untrained RNN with randomly-initialized reservoirs, the particular ESN/nVAR models used here are taken from prior works that introduce additional architectural modifications like imposed sparsity and ridge regularization \cite{gauthier2021next}. These domain-specific modifications allow these models to outperform trained RNN on forecasting tasks, despite their architectural similarities.

While the utility of deep learning methods for forecasting general time series has been questioned \cite{makridakis2022m5,godahewa2021monash}, our results agree with recent benchmarks suggesting that large models strongly outperform classical forecasting methods on long-horizon forecasting tasks \cite{zhou2021informer}. We find that classical methods like exponential smoothing or ARIMA do not appear among the top models, implying that the size and diversity of our chaotic systems dataset, as well as the long duration of the forecasting task, require larger models with greater intrinsic capacity to represent complex nonlinear systems. Relative performance among models remains stable across two orders of magnitude in Lyapunov time, indicating that strong models better approximate the underlying propagator for the flow even at small forecasting horizons. Given the autoregressive nature of forecasting, an initial accuracy advantage compounds over time due to the exponential sensitivity of chaotic systems to early errors. However, in the supplementary material we show that the best-performing models \textit{also} reproduce dynamical invariants such as Lyapunov exponent spectra and fractal dimensions better than other methods, suggesting that pointwise forecast accuracy is a prerequisite to accurately reconstructing dynamical manifolds.

\subsection{The inductive biases of physics-based models provide advantages in data or compute-limited settings}. 

\begin{figure*}[!]
{
\centering
\includegraphics[width=0.7\linewidth]{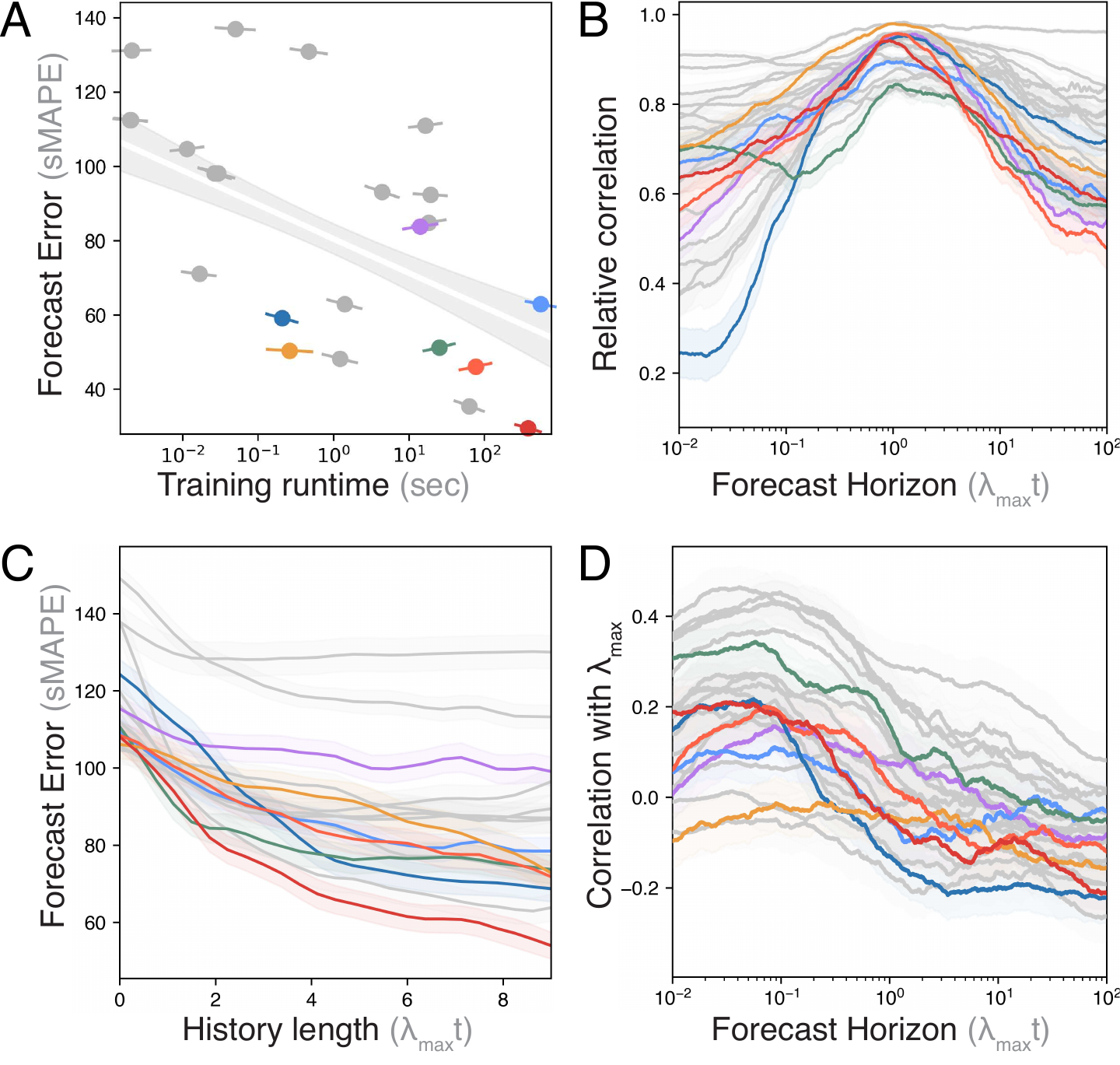}
\caption{
{\bf Universal relationships among forecasting methods.} (A) Error versus training time at fixed forecast horizon $t = \lambda_\text{max}^{-1}$ for all models. Bar lengths denote standard deviations along principal axes, with angle indicating Spearman correlation within each model group in order to detect Simpson's paradox. The underlaid linear fit indicates the overall correlation $\rho = -0.31 \pm 0.04$. (B) Median relative correlation of each forecasting method with its average prediction, across different forecast horizons. (C) Median model errors at $t = \lambda_\text{max}^{-1}$ as the amount of history data increases. (D) Correlation of forecasting error with Lyapunov exponent $\lambda_\text{max}$ as a function of forecasting horizon. All error bars correspond to 95\% confidence intervals, and colors match methods from previous figures.
}
\label{relation}
}
\end{figure*}

While domain-agnostic time series methods perform well overall, we note that the different forecasting methods have different intrinsic model complexities and thus capacities. Fig. \ref{relation}A shows the forecasting error at $\lambda_\text{max}^{-1}$ versus the computational walltime required to train each model on one central processing unit. Training walltime measures model efficiency, and we interpret it as a loose proxy for model complexity because model size and trainable parameter count are not directly quantifiable across highly-distinct and regularized architectures \cite{canziani2016analysis}. We find that error and training time exhibit negative correlation ($\rho = -0.31 \pm 0.04$, bootstrapped Spearman coefficient), which persists within most method groups. The best-performing machine learning models require considerable training times; in contrast, reservoir computers (both nVAR and ESN) exhibit competitive performance with two orders of magnitude less training time due to their linear structure. The strong performance of reservoir computers implicates an inductive bias for learning complex dynamical systems, due to their fixed kernel structure allowing them to more readily represent continuous spectra \cite{jaeger2004harnessing,kim2021teaching,smith2022learning,sussillo2009generating}. In particular, the nVAR model regresses a set of fixed nonlinearities that includes quadratic terms, making it possible for this model to learn an exact multistep integration scheme for many models in our dataset \cite{zhang2023catch}. In contrast, the transformer model has high intrinsic capacity and likely a low inductive bias for dynamical systems \cite{zhou2021informer}, and thus requires the most computational resources.

\subsection{Invariant properties fail to explain long-term predictability of different chaotic systems}. 

This general tradeoff between performance and training difficulty motivates us to search for universal similarities across different forecasting methods. In Fig. \ref{relation}B we compute the Spearman correlation between each method's instantaneous and time-averaged forecast error as a function of the forecast horizon, $\rho_k(t)$ where $k$ indexes the forecast method. We find universal non-monotonic behavior, in which nearly all methods exhibit peak correlation at one Lyapunov time $\lambda_\text{max}^{-1}$ (Fig \ref{relation}B). This observation underscores that the largest Lyapunov exponent represents an appropriate timescale for comparing different dynamical systems, and that diverse forecasting models interact with this property in a shared manner. $\lambda_\text{max}^{-1}$ is sufficiently long to distinguish dynamical systems based on their invariant properties, but short enough that forecast methods do not accrue instabilities, large phase offsets, and other artifacts that saturate forecast error and mask intrinsic differences among systems. This observation aligns with recent work from statistical learning theory that draws analogies between trained learning models and disordered systems \cite{martin2021predicting}: when models become most strongly coupled to the specific properties of individual systems, they exhibit peak correlation with their mean-field prediction. Additionally, we find the predictions of different large models become correlated at long forecasting horizons, suggesting that they agree on which particular dynamical systems prove hardest to forecast (Fig. \ref{relation}D). Surprisingly, their performance only weakly correlates with $\lambda_\text{max}^{-1}$ and thus intrinsic chaoticity, with any correlation vanishing at long forecasting horizons. This unintuitive finding suggests either (a) that large forecasting models have not yet reached sufficient scale and refinement that their performance is bounded by $\lambda_\text{max}$; or (b) that $\lambda_\text{max}$ is not the only invariant quantity that governs the empirical long-term predictability of a dynamical system.

Taken together, these results suggest that the greater intrinsic capacity and flexibility of overparameterized domain-agnostic models allows them to access a new long-horizon forecasting regime, in which their forecast accuracy decorrelates with Lyapunov exponent---and thus intrinsic chaoticity. To further investigate model complexity and performance, we next perform a series of experiments in which we titrate the history length $t^*$, which determines the total amount of training data available to each method before generating a forecast (Fig. \ref{relation}C). Unlike training time or parameter count, this quantity determines how effectively different methods utilize additional observations of a chaotic attractor. As expected, all models asymptotically improve given additional training data. However, while the neural ordinary differential equation and nVAR models both exhibit initially steep drops---indicating favorable performance in the low-data regime---they fail to reach asymptotic errors lower than the larger-scale models that dominate when more data is available. However, NBEATS performs well in both the low-data and asymptotic regimes, suggesting that the neural basis expansions used in its earlier stages provide an inductive bias that dominates when less training data is available. This matches the intuition that performant models should require both high intrinsic capacity and inductive biases for dynamical systems.

%%%%%%%%%%%%%%%%%%%%%%%%
%
\section{Discussion}
%
%%%%%%%%%%%%%%%%%%%%%%%%

Our results show that recently-developed large, overparameterized statistical forecasting models efficiently leverage long-term observations of chaotic attractors, producing best-in-class forecasts that can remain accurate for up to two dozen Lyapunov times. Commonalities in predictions across highly distinct model classes suggest that performance arises primarily from model capacity and generalization ability, rather than specific architectural choices, and that performance at long prediction times is ultimately limited by a model's ability to learn long-term properties of a dynamical system's underlying attractor. The strong performance of generic large models echoes recent findings from other domains, and it represents an intuitive consequence of the "no free lunch" theorem for model selection \cite{brown2020language,wolpert1997no}. Nonetheless, our results are practically informative for forecasting real-world time series driven by underlying dynamical systems. In the absence of restrictions on data availability or training resources, large domain-agnostic models are likely to produce high-quality forecasts without the need for system-specific knowledge. However, in restricted settings, domain-specific methods such as reservoir computers exhibit the strongest performance relative to their computational requirements \cite{gauthier2021next}.
%cite sutton

While certain methods perform particularly well in our experiments, we refrain from endorsing specific models to the detriment of others: our results may be specific to our chaotic systems dataset and, more importantly, the recent literature contains a broad variety of new forecasting models, as well as infinite possible variations of each method due to hyperparameter and architectural choices, which could potentially exhibit comparable performance. Rather, we have chosen a representative set of forecasting models bridging different foci of the literature \cite{makridakis2022m5,lim2021time}, and highlight general trends and the emerging strength of new models on the classical problem of forecasting chaos.

Our observation that $\lambda_\text{max}$ fails to fully determine whether a system remains empirically predictable over extended horizons introduces the possibility that our empirical forecasting results may instead correlate with other invariant properties of the different dynamical systems in our dataset, such as various measures of fractality and entropy \cite{hunt2015defining,tang2020introduction}, or covariant Lyapunov spectra \cite{margazoglou2023stability,cvitanovic2005chaos,ozalp2023reconstruction}. Such characterization could improve the interpretability of machine learning-based forecasting models, which ostensibly provide less insight into a time series's structure than classical methods \cite{oreshkin2020nbeats,godahewa2021monash}. However, the strong empirical performance of machine learning suggests the potential for these methods to reveal new properties of nonlinear dynamics and, ultimately, new bounds on the intrinsic predictability and thus reducibility of chaotic systems.

\section{Code availability}

All code used in this study is available online at \url{https://github.com/williamgilpin/dysts}

\section{Acknowledgments}

We thank E. L. Florin and Yuanzhao Zhang for feedback on the manuscript. Computational resources for this study were provided by the Texas Advanced Computing Center (TACC) at The University of Texas at Austin. This project has been made possible in part by grant number DAF2023-329596 from the Chan Zuckerberg Initiative DAF, an advised fund of Silicon Valley Community Foundation.

%\clearpage

\renewcommand{\thetable}{S\arabic{table}}
\setcounter{table}{0}
\renewcommand{\thefigure}{S\arabic{figure}} 
\setcounter{figure}{0}
\renewcommand{\theequation}{A\arabic{equation}}
\setcounter{equation}{0}
\renewcommand{\thesubsection}{\Alph{subsection}}
\setcounter{subsection}{0}
\renewcommand{\thesection}{Appendix \Alph{section}}
\setcounter{section}{0}

\section{Example Forecast Trajectories}

\begin{figure}
{
\centering
\includegraphics[width=\linewidth]{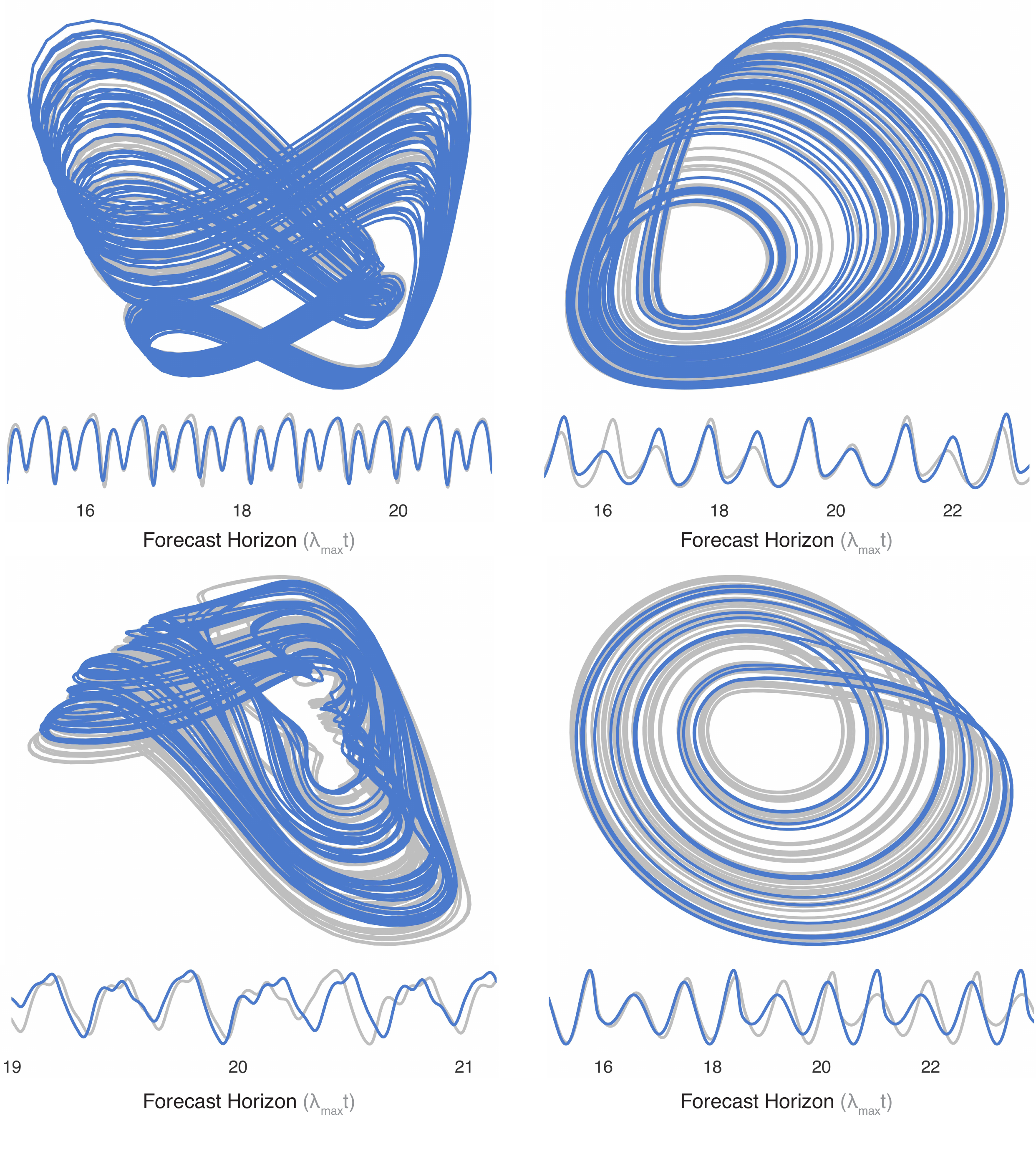}
\caption{
\textbf{The most predictable systems in the dataset.} Long-term forecasts for a range of systems in the dataset for which accurate forecasts are achieved for $>20 \lambda_\text{max}^{-1}$. The predictions (blue) correspond to the best-performing forecast model on that particular system, relative to a held-out true trajectory (gray). Predictions versus time for a single dynamical variable are underlaid, in order to emphasize pointwise accuracy. Each forecast is generated fully autoregressively, receiving initial conditions and their preceding values but no other input.
}
\label{long_forecast}
}
\end{figure}

To complement our quantitative results, in Figure \ref{long_forecast} we show particular examples of the most accurate forecasts at $>20 \lambda_\text{max}^{-1}$ for several attractors in our dataset. In all examples, the best forecast is produced by NBEATS/NHiTS, and the predictions both matche the global attractor geometry and point-to-point variations in the dynamics. We note that these particular forecasts represent especially predictable systems from within our dataset; however, if we define a prediction time based on the latest time when $\text{SMAPE} < 50$, the average valid forecast horizon of the best-performing model averaged across all $135$ systems is equal to $13.9 \pm 2.8 \,\lambda_\text{max}^{-1}$. If we instead define the prediction time based on the first doubling of accumulated pointwise errors, we find a horizon of $15 \pm 5.3 \,\lambda_\text{max}^{-1}$.

\section{The chaotic systems dataset.}
\label{a_dataset}

\begin{figure}
{
\centering
\includegraphics[width=\linewidth]{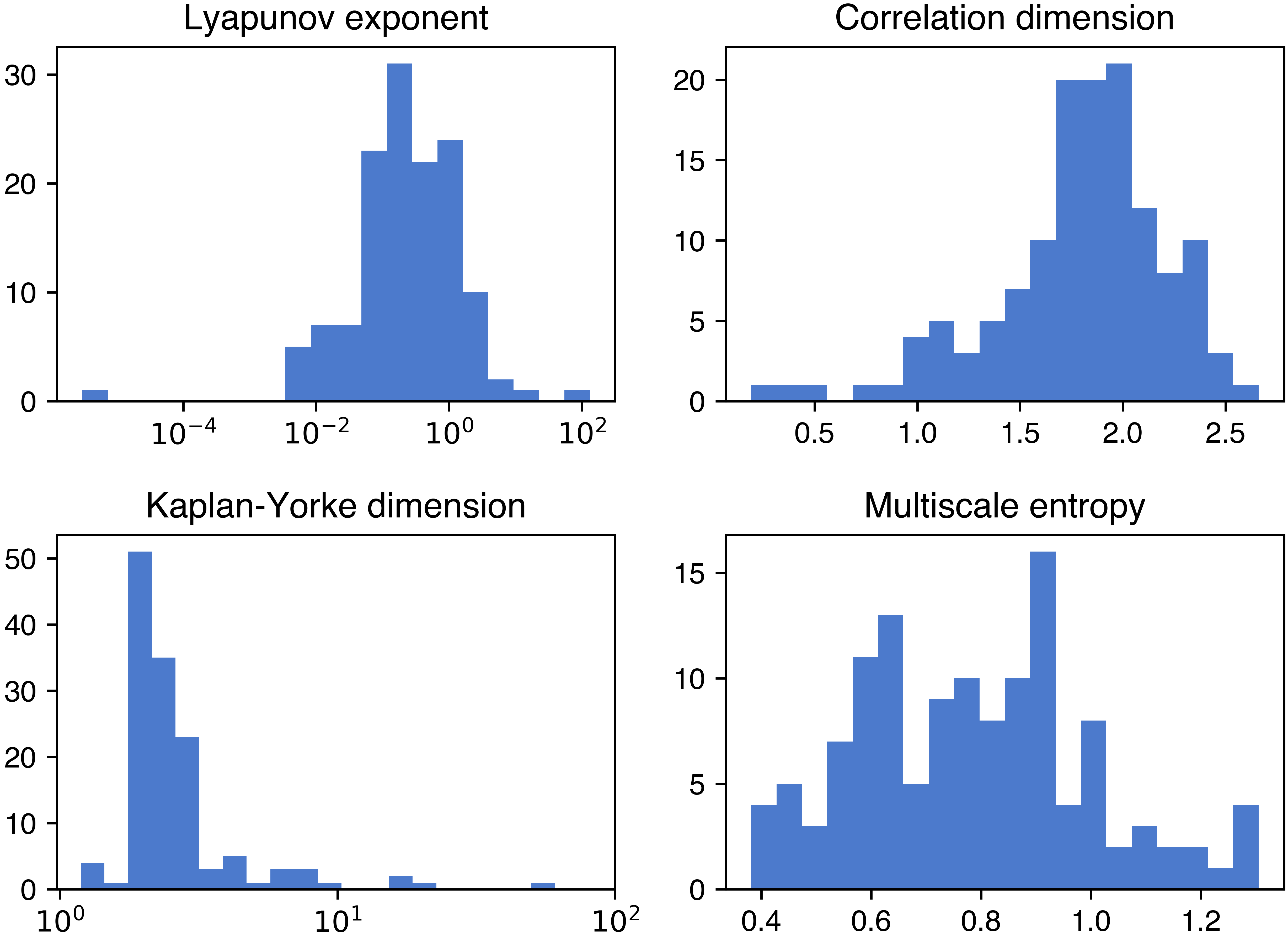}
\caption{
\textbf{Invariant properties of the dynamical systems dataset.} Histograms showing the number of systems (out of $135$ total) with invariant properties in each bin range. Across the systems, the maximum Lyapunov exponent $\lambda_\text{max} = 0.24 \pm 0.31$, the correlation dimension $D_2 = 1.84 \pm 0.22$, the Kaplan-Yorke dimension $D_\text{KY} = 2.19 \pm 0.45$, and the multiscale entropy $E = 0.78 \pm 0.15$.
}
\label{stats}
}
\end{figure}

Our dynamical systems dataset corresponds to an expanded version of our initial benchmark \cite{gilpin2021chaos}; after the release of our initial benchmark, additional systems were suggested and submitted by users of the open-source code. Each dynamical system in our dataset has the form $\dot{\v{x}} = \v{f}_k(\v{x})$, $k \in \{1, 2, ..., 135\}$, and non-autonomous systems are lifted by defining a time-like dynamical variable. For each dynamical system we compute the maximum Lyapunov exponent $\lambda_\text{max}$, correlation fractal dimension, Kaplan-Yorke fractal dimension, and the multivariate multiscale entropy. Figure \ref{stats} shows distributions of each quantity across the $135$ systems.

All systems in our database are timescale-aligned to have matching dominant timescales and sampling rates: for each system, we calculate the optimal integration timestep by computing the power spectrum, and then using random phase surrogates to identify the smallest significant frequency $1 / t_{max}$ (a lower bound on the Lipschitz constant) and the dominant significant frequency $1 / t_{peak}$ \cite{kantz2004nonlinear}. The smallest frequency determines the integration timestep  $t_{max} / 10$ for all numerical integration. In order to ensure that trajectories are timescale-aligned, after integration trajectories are resampled to $100$ timepoints per $t_{peak}$. Unless otherwise noted, we measure natural time in units of the dominant significant Fourier timescale $t_{peak}$, though we rescale this quantity by the Lyapunov exponent when reporting results in the main text. We explore dependence of forecasting on time series granularity (sampling rate) and added stochasticity in previous work \cite{gilpin2021chaos}; here we focus on a fixed fine-granularity consisting of trajectories with $100$ timepoints per dominant Fourier period $t_\text{peak}$.

Properties of our dataset, invariant property calculation, and system selection procedure are described in detail in prior work \cite{gilpin2021chaos}, and all code used to prepare and analyze our dataset is included in our open-source code.

\section{Forecasting experiment design.} 

Having identified natural units of time for each system in terms of $t_\text{peak}$, we next define the structure of our forecasting task in terms of these units. There are several timescales for our forecasting task:
\begin{enumerate}
\item \textbf{Total trajectory length} $T$. This corresponds to the total length of the trajectory computed for separate initial conditions for the train and test sets. This is the longest timescale used in our experiments, and it is later subdivided into history (for training model parameters) and the forecast horizon.
\item \textbf{Total trajectory index} $t'$. This time satisfies $t' \in [0, T]$, and it measures absolute time along the original trajectory. We use this notation in order to reserve the quantity $t$ to refer to forecast horizon (time since training data is no longer available).
\item \textbf{Lookback window} $T_\ell$. The number of datapoints seen simultaneously by a forecast method at any given time during both training and forecasting. This quantity is akin to the number of features seen simultaneously by a linear regression or random forest model, or the number of lags in autoregressive state space models. Physically, $T_\ell$ represents the amount of context information from past values that a fully-trained model can access when generating a forecast.
\item \textbf{History length} $t^*$. This corresponds to the total amount of training data $t^* \geq T_\ell$, or the total number of unique past values seen by the model across all training iterations and epochs. For our experiments, $t^*$ is equal to $10$ periods of $t_\text{peak}$.
\item \textbf{Forecast horizon} $t$. The number of timepoints into the future that a model forecasts after training on $[0, t^*)$. In discrete time, $t = 1$ represents the next timepoint immediately after timepoints $1, 2, ..., t^*$ have elapsed. Note that $t$ is defined as the time since the end of training data availability, $t \equiv t' - t^*$. For our experiments, $t \in (0, T - t^*]$, which is equivalent to $t' \equiv (t^*, T]$.
\item \textbf{The Lyapunov time} $\lambda_\text{max}^{-1}$. An invariant property distinct to each chaotic system, representing the characteristic timescale over which forecasts are expected to lose accuracy due to the butterfly effect. We scale many of our forecast results to this time, $\lambda_\text{max} t$.
\end{enumerate}

For all systems, we use a fixed duration of $t^* = 10\, t_\text{peak}$ for all training data. For validation, model selection, and hyperparameter tuning, we generate forecasts up to an additional $t = 2 \, t_\text{peak}$ after $t^*$. For evaluating performance, we generate predictions and timepoint-wise forecast error metrics for all future horizons spanning $t \in [0.01 - 50]$ $t_\text{peak}$, thus allowing us to calculate a horizon-dependent error score $\epsilon_{ik}(t)$. This testing dataset corresponds to a trajectory generated from different initial conditions than the trajectory used for training and validation (model selection).

\section{Forecast models evaluated.} 

We choose $24$ forecasting methods spanning a variety of areas of the forecasting and dynamical systems literature, including state-of-the-art methods \cite{oreshkin2020nbeats,lea2016temporal,alexandrov2020gluonts,godahewa2021monash}. Our forecasting methods can be grouped into several categories:

\subsection{Physics-based methods.} These models contain inductive biases for time series that would give them an advantage when forecasting time series generated by dynamical systems.
\begin{itemize}
	\item \textbf{Echo state networks (ESN).} Our echo-state network implementations represent standard configurations used in recent works \cite{vlachas2020backpropagation,tanaka2019recent}. We note that many variants of echo state networks and reservoir computers exist, which use different nonlinear activation functions, reservoir sizes and initializations, and other structural features. Much like architecture choices for deep learning models, it is infeasible to consider the space of all possible models, and so we default to standard architectural choices used in prior works. This includes a fixed reservoir size of $500$ units, spectral radius of $0.99$, reservoir connectivity of $0.1$, input scaling of $1.0$, input connectivity of $0.2$, and a ridge regularizer of strength $10^{-4}$ in the readout layer. During model selection, we evaluate the leakage rate hyperparameter across the range $0.01$ to $1.2$.

	\item \textbf{Nonlinear Vector Autoregression (nVAR).}  This model uses a single hidden layer to produce nonlinear combinations of the input features, which correspond to past time points. Regularized linear regression on these lifted features is used to generate forecasts. Recent work has shown that these models are equivalent to reservoir computers given a sufficient number and diversity of nonlinear features \cite{gauthier2021next,bollt2021explaining}. Following prior works, we use default hyperparameter values, including a fixed reservoir delay of $100$ ($1$ $\lambda_{max}^{-1}$ in our units), and apply a ridge regularizer of strength $10^{-4}$ in the readout layer. During model selection, we evaluate the leakage rate parameter in the range $0.01$ and $1.2$.
	
	\item \textbf{Neural Ordinary Differential Equations (nODE).} These models use deep neural networks to represent learn the function $\v{f}$ in an equation $\dot{\v{x}}(t) = \v{f}(\v{x}, t)$ \cite{chen2018neural}. The neural network takes in the initial state $\v{x}(t_0)$ and produces the trajectory of $\v{x}(t)$ over time via numerical integration. The network is trained by comparing the integrated trajectory with the true trajectory, and using the error to update the function's parameters using either the adjoint method or backpropagation---which become equivalent in continuous time \cite{sokol2019adjoint}. We use default hyperparameters, corresponding to a two-layer $30$ unit residual network with SiLU activation, trained for $500$ epochs with a learning rate of $0.01$ and a batch size of $128$ samples. During model selection, we provide the network a varying amount of previous timepoints, corresponding to tuning the lookback window hyperparameter across the range $2 -120$.
\end{itemize}
	
\subsection{Deep learning methods.} These methods represent large models with many trainable parameters, which are usually trained iteratively using variants of gradient descent. The term "deep" traditionally refers to models with many trainable layers, though here we use it more informally to refer to overparameterized models with hierarchical structure, in contrast to classical machine learning methods.
\begin{itemize}
	\item \textbf{NBEATS}. NBEATS (Neural basis expansion analysis for interpretable time series forecasting) is an artificial neural network architecture that uses a stack of fully-connected layers, and residual connections among layers, to model the past and future values of a time series \cite{oreshkin2020nbeats}. It does not rely on any time-series-specific components such as recurrent or convolutional layers, and can produce interpretable outputs by using either pre-specified or fully-trainable functions. Here, we use fully-trainable basis functions, in order to avoid making assumptions about the structure of the training data. We use default hyperparameters from a reference implementation, corresponding to $4$ layers each containing $256$ units and ReLU activation functions. During model selection, we evaluate the input length parameter in the range $2 - 50$ timepoints.
	\item \textbf{NHiTS.} A model that builds upon NBEATS by using hierarchical interpolation and multi-rate input processing to specialize its predictions for different significant frequencies in the input signal \cite{challu2023nhits}. NHiTS consists of several fully-connected blocks with residual connections, which operate on downsampled versions of the time series before upsampling their forecasts back to the input shape. NHiTS has been shown to match or exceed the performance of NBEATS on standard reference datasets, while substantially reducing required computational resources. We use default hyperparameters from a reference implementation, corresponding to $2$ layers each containing $256$ units and ReLU activation functions. During model selection, we evaluate the input length parameter in the range $2 - 50$ timepoints.
	\item \textbf{Transformer.} A type of artificial neural network architecture that use self-attention mechanisms to process sequential data, such as natural language or time series \cite{vaswani2017attention}. Transformers can capture long-term dependencies and complex patterns in time series data by using positional encoding and multi-head attention, leading to their widespread use for diverse problems such as machine translation, text summarization, and question answering \cite{brown2020language}. We use an architecture based on the Informer, a model recently shown to exhibit state-of-the-art performance on long-duration forecasting tasks \cite{zhou2021informer}. We use default hyperparameter values from a reference implementation corresponding to $4$ attention heads, $3$ encoder layers, and $512$ nodes in the feedforward layers. During model selection, we evaluate the input length parameter in the range $2 - 50$ timepoints.
	\item \textbf{RNN.} A neural network architecture that sequentially updates a hidden state based on a combination of the hidden state's previous value, and new inputs. RNN models are widely-used for sequential data, and represent a starting point for more recent, specialized architectures for time series and natural language processing. We use default hyperparameter values from a reference implementation containing $2$ recurrent layers with $25$ units. During model selection, we evaluate the input length parameter in the range $2 - 50$ timepoints.
	\item \textbf{LSTM.} A type of recurrent neural network with specialized gating architecture that better allows incorporation of long-range information \cite{hochreiter1997long}. The architecture prevents vanishing gradients during training, leading to strong performance on data assimilation, time series representation, and language modelling tasks. We use default hyperparameter values from a reference implementation containing $2$ recurrent layers with $25$ units. During model selection, we evaluate the input length parameter in the range $2 - 50$ timepoints.
	\item \textbf{Temporal Convolutional Network.} A neural network architecture that uses one-dimensional convolutional layers with causal connections to capture the temporal dependencies \cite{lea2017temporal}. Unlike a traditional convolutional neural network, convolutions are strided to ensure that predictions only depend on past values of the time series. Our TCN consists of several stacks of dilated convolutional layers with residual skip connections that increase the receptive field while preserving the input length. The same architecture recently achieved best-in-class performance for unsupervised time series featurization \cite{franceschi2019unsupervised}. We use default hyperparameter values from a reference implementation with a dilation factor of $2$ and $3$ convolutional filters of width $3$ \cite{bai2018empirical}. During model selection, we evaluate the input length parameter in the range $2 - 50$ timepoints.
%	\item \textbf{Neural Ordinary Differential Equations (nODE)}. Neural ODEs use deep neural networks to learn a function $\v{f}$ representing a dynamical system $\dot{\v{x}}(t) = \v{f}(\v{x}, t)$, which reproduces the training time series when numerically integrated \cite{chen2018neural}. These methods exploit an equivalence between backpropagation algorithms traditionally used to train deep neural networks, and the adjoint method commonly used to determine parameter values for differential equations.
\end{itemize}

\subsection{Modified linear models.} These recently-proposed linear models represent ablations isolating different properties of the Transformer architecture \cite{zeng2022transformers}, in order to identify which aspects of the architecture most strongly determine its performance on a given dataset.
\begin{itemize}
	\item \textbf{DLinear.} A model that decomposes a time series into its leading trend component via a moving average, and a residual seasonal component. It then combines these components to produce a forecast. The original authors expect this model to perform more strongly when the data has a strong trend component. During model selection, we evaluate the input size hyperparameter in the range of $2 - 50$ previous timepoints.
	\item \textbf{NLinear.} A linear model that attempts to account for distribution shift or non-ergodicity (or apparent non-ergodicity due to insufficient sampling). Trailing points from the time series history used for training are used to establish a baseline, which is first removed before fitting, and then added back to the forecast. During model selection, we evaluate the input size hyperparameter in the range of $2 - 50$ previous timepoints.
\end{itemize}

\subsection{Classical Statistical Methods.} These methods represent common statistical forecasting techniques, which are widely-used but which are not overparameterized relative to the training dataset size.
\begin{itemize}
	\item \textbf{ARIMA}. Autoregressive Integrated Moving Average (ARIMA) is a linear time series forecasting model that combines autoregression (AR), differencing (I), and moving average (MA) components to capture various patterns in the data, such as trends and seasonality \cite{hyndman2018forecasting}. Specified by three hyperparameters $(p, d, q)$, the ARIMA model is effective in situations where the underlying data-generating process can be well-approximated by a linear combination of past values and errors. 
	\item \textbf{AutoARIMA}. A variant of the ARIMA model that enforces stationarity by differencing until the data no longer rejects the null hypothesis of stationarity under the Kwiatkowski-Phillips-Schmidt-Shin (KPSS) test, and which then automatically determines model order based on the Akaike Information Criterion $\text{AIC} = 2 k - 2 \log\text{Err}$, which penalizes larger models (with $k$ parameters) that fail to decrease the training fit error $\text{Err}$. During model selection, we evaluate the lag order hyperparameter in the range of $2 - 50$ previous timepoints. 
	\item \textbf{Exponential Smoothing}. A family of forecasting methods that apply exponentially decreasing weights to past observations, with more recent observations receiving higher weights \cite{hyndman2018forecasting,gardner1985exponential}.
	\item \textbf{Theta.} A univariate forecasting technique that transforms the time series into two new time series in which short-term and long-term fluctuations are amplified, respectively \cite{assimakopoulos2000theta}. The two modified time series are separately forecast using exponential smoothing, and the resulting predictions are then combined to generate a final forecast. We use a fixed theta value equal to $2$, in order to differentiate the model from pure exponential smoothing.
	\item \textbf{Four Theta.} An extension of the Theta method, which performs additional smoothing and amplification transforms in order to isolate fluctuations over a greater range of timescales.  We use a fixed theta value equal to $2$, in order to differentiate the model from pure exponential smoothing.
\end{itemize}

\subsection{Classical Machine learning.} These represent common regression methods used in machine learning, which are not overparameterized relative to the training dataset size.
\begin{itemize}
	\item \textbf{Linear Regression}. A standard linear regression between past values of the time series and the next value. A weak, untuned ridge regularization term of amplitude $0.01$ prevents the model's weights from diverging during training. During model selection, we evaluate the input size hyperparameter in the range of $2 - 50$ previous timepoints.
	\item \textbf{Fourier Transform Regression}. A set of dominant frequencies and relative phases are identified from a time series' power spectrum. Forecasts are generated by repeating these frequency components indefinitely into the future, with an appropriate phase offset.
	\item \textbf{Random Forest}. A model that trains a series of decision tree regressors on individual subsets of the timepoints in the lookback window, and then averages their predictions to produce a consensus forecast \cite{hastie2009elements}. In contrast, the gradient-boosting model XGBoost trains an ensemble of trees sequentially, such that each tree prioritizes forecasting timepoints on which previous trees underperformed \cite{friedman2001greedy}. We use default hyperparameters consisting of $100$ trees, with each tree's depth allowed to grow until either all leaves are pure, or all leaves contain less than $2$ samples. During model selection, we evaluate the number of input features hyperparameter in the range of $2 - 50$ previous timepoints.
	\item \textbf{XGBoost}.  A variant of the Random Forest, in which individual decision trees are trained sequentially to improve on earlier trees' outputs. XGBoost approaches state-of-the-art performance on regression and classification of tabular data \cite{grinsztajn2022tree}. Surprisingly, recent benchmarks suggest that XGBoost can outperform artificial neural networks on time series forecasting \cite{elsayed2021we}, despite the model lacking structural inductive biases that exploit temporal correlations and continuity. During model selection, we evaluate the number of input features hyperparameter in the range of $2 - 50$ previous timepoints.
\end{itemize}

\subsection{Naive baselines.} These models isolate single properties of the training time series, and use it to generate minimal forecasts based on solely that attribute. Much like ablations used to evaluate machine learning models \cite{sheikholeslami2021autoablation}, these baselines provide minimal reference values against which to compare the results of more sophisticated methods.
\begin{itemize}
	\item \textbf{Naive Mean.} A model that forecasts all future values of the time series as equal to the mean of the previous values. 
	\item \textbf{Naive Drift.} A model that extracts the dominant linear trend from the previous values, and extrapolates that trend into the future.
	\item \textbf{Naive Seasonal.} A model that determines the dominant phase and timescale using the peak of the power spectrum, and then computes the average of a set of non-overlapping consecutive windows with width equal to the dominant timescale (after first applying the phase shift). The model then continues this repeated motif indefinitely to generate a singly-periodic forecast of future values.
	\item \textbf{Unforced Kalman}. A recursive Bayesian estimator that uses a set of linear equations to optimally estimate the state of a dynamic system, given noisy and partial observations \cite{kantz2004nonlinear}. The Kalman filter recursively updates its state estimate by predicting the next state using the state-transition model and refining the estimate with new observations via the observation model. In a forecasting context, the transition matrix and other parameters are fit using the training data. To generate a forecast after the training history ends, the internal model state is held constant without updating in order to propagate the forecast autoregressively without additional input data. This naive approach provides a lower-bound on the performance of linear models directly fit on the training data \cite{hyndman2018forecasting,harvey1990forecasting,durbin2012time}.
\end{itemize}

For most deep learning models, we use reference implementations provided by the \texttt{darts} Python library \cite{herzen2022darts}.  For other models, we use reference implementations in the \texttt{statsmodels}, \texttt{Gluon-TS}, \texttt{sktime}, and \texttt{scikit-learn} libraries. We use the authors' original codes for the neural ODE and reservoir computer models \cite{loning2019sktime,alexandrov2020gluonts,chen2018neural}.
All untuned hyperparameters (e.g. batch size, training epochs, model width, number of layers, etc) are kept at default values used in reference implementations.

\section{Hyperparameter tuning, validation, and model selection.} 

We tune hyperparameters separately for each forecasting model and dynamical system pair. For each trajectory, $10$ full periods comprising $1000$ timepoints are used to train the model, and $2$ additional periods comprising $200$ timepoints are used to estimate sMAPE errors for each combination of hyperparameters. Measured in terms of Lyapunov times $\lambda_\text{max}^{-1}$, this corresponds to an average of $11 \,\lambda_\text{max}^{-1}$ used for training each system.

Though distinct forecasting methods have different hyperparameters and architectural details, we focus on tuning whichever hyperparameter most closely corresponds to the {\it lookback window} $T_\ell$ for each method. This corresponds to the number of timepoints that the model sees simultaneously when generating a prediction for the next timepoint. In the context of the specific methods considered here, it corresponds to the "lag order" of traditional auto-regressive models like ARIMA. For classical machine learning methods and artificial neural networks, this corresponds to the "input size," or the number of features seen by the model simultaneously as input. For reservoir computers, it corresponds to the leakage rate, which indirectly influences the reservoir's spectral radius \cite{lukovsevivcius2012practical}.

We treat classical models accepting a seasonality hyperparameter as multiple distinct models, and select the best-performing one. A standard grid search via time series cross validation on the training data determines the optimal hyperparameters separately for each method and dynamical system pair.

\textbf{Caveats of model selection and justification.}  Complex forecasting models contain many hyperparameters and architectural choices, which parameterize an infinite set of possible models related to each specific method that we test. For example, deep neural networks offer many choices regarding number of layers, depth, learning rate, and batch size, while reservoir computers require choices regarding the random initialization scheme for the reservoir, reservoir size, and unit dynamics. In the spirit of previous large-scale benchmarks \cite{schmidt2021descending,olson2017pmlb}, we seek to perform comparable degrees of model selection for each of the $24$ forecasting models that we consider. Thus the best-performing methods represent those that achieved strongest performance given the particular hyperparameters and value ranges we consider, and do not necessarily preclude other models from exhibiting comparable performance in certain regimes. Nonetheless, our results suggest that certain forecasting methods lend themselves to producing strong results with minimal fine-tuning.

\section{Accuracy metrics}
\label{a_results}

We consider $16$ different forecast accuracy metrics, though we report our main text results in terms of one metric, sMAPE, due to its widespread use, favorable properties, and interpretability \cite{godahewa2021monash,makridakis2022m5,gilpin2023recurrences,gilpin2021chaos,wang2023koopman}. We include all other results in tabular form in our open-source code repository.

\subsection{Pointwise metrics}

\begin{figure*}
{
\centering
\includegraphics[width=\linewidth]{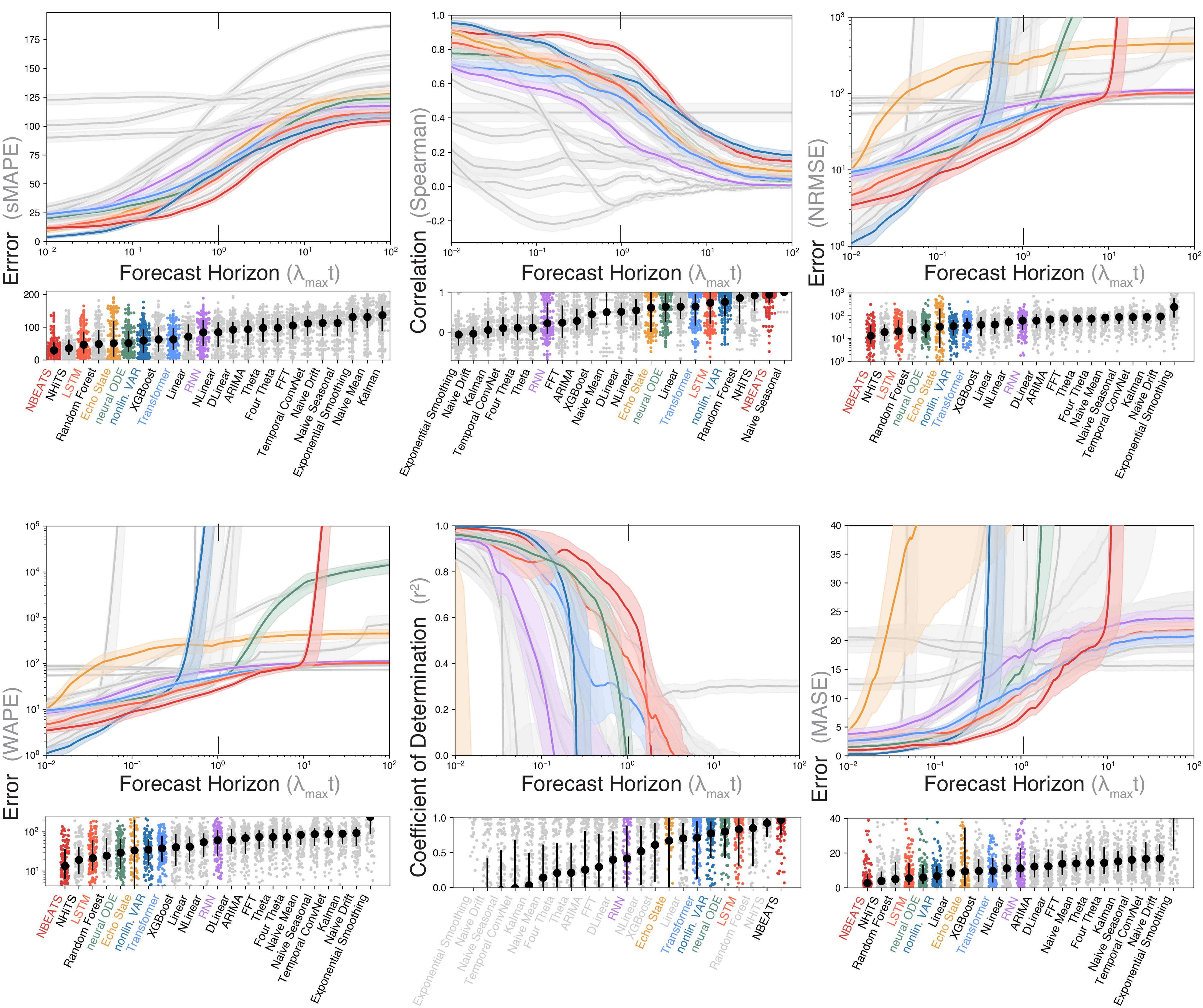}
\caption{
Alternative error metrics for the forecasting experiments. Many normalized metrics exhibit sensitivity to fluctuations in their denominators, resulting in values exceeding the axes bounds as the errors diverge. Colors correspond to highlighted models from the main text.
}
\label{points}
}
\end{figure*}

A summary of our results in terms of various point-wise accuracy metrics is shown in Figure \ref{points}, which pairs the accuracy of each method over gradually increasing forecasting horizons $t$ with a snapshot of the distribution of scores at one Lyapunov time $\lambda_\text{max}^{-1}$. Here, we describe each error metric in terms of a true trajectory $\v{y}(t')$ and predicted trajectory $\hat{\v{y}}(t')$. We suppress the subscripts $i$ and $k$, which we use elsewhere to denote the performance of the $i^{th}$ forecasting model on the $k^{th}$ dynamical system.

\begin{enumerate}
\item \textbf{Symmetric mean absolute percent error.} The sMAPE scales the absolute percent error based on the magnitude of the two input time series. If $\v{y}(t')$ is the true trajectory and $\hat{\v{y}}(t')$ is a predicted trajectory sampled at a discrete set of values $i' \in 1, 2, ... t$, then the sMAPE is defined as
\[
\epsilon(t) \equiv \dfrac{200}{t} \sum_{t'=1}^t \dfrac{| \v{y}(t') - \hat{\v{y}}(t')|}{|\v{y}(t')| + |\hat{\v{y}}(t')|},
\]
This metric is widely-used in prior forecasting studies due to its compact range and interpretability \cite{godahewa2021monash,makridakis2022m5,gilpin2023recurrences,gilpin2021chaos,wang2023koopman}. The argument $t$ indicates that this instantaneous error signal depends on how far into the future a forecast is generated---the forecast horizon. When referring to the error associated with a specific forecasting method on a particular dynamical system, we use subscripts $\epsilon_{ik}(t)$, with $i$ indexing the forecast method and $k$ indexing the dynamical system.

\item \textbf{Spearman correlation.} This metric measures the tendency of the true and forecasted time series to co-vary, independently of the relative magnitude of either series. We find that this metric performs well and cleanly differentiates the models, though it tends to zero over extended periods as the true time series and the forecast decorrelate. The relative ordering of different forecasting models remains largely the same under this metric, though the nVAR model gains a slight relative advantage. 

\item \textbf{Normalized Root Mean Squared Error.} The NRMSE rescales the RMSE relative to the average fluctuations within the time series,
\[
\sqrt{ \dfrac{1}{t \; d}\sum_{t'=1}^t \left(\dfrac{(\v{y}(t') - \hat{\v{y}}(t'))^\top (\v{y}(t') - \hat{\v{y}}(t'))}{\sigma^2} \right) }
\]
where $\v{y}, \hat{\v{y}} \in \mathbb{R}^{d}$ and $\sigma$ corresponds to the standard deviation of $\v{y}(t')$. There is some ambiguity regarding the calculation of $\sigma$, which can be estimated from the full training data, the forecast interval, or external measurements. In order to mitigate sampling errors in the calculation of $\sigma$ due to uneven forecast horizons, we estimate $\sigma$ over the entire training dataset, making it a constant for each dynamical system. This metric is prone to divergence, like the other metrics considered here. Nonetheless, the relative ordering of the models remains consistent, though the neural ODE model fares slightly better under this metric.

\item \textbf{Mean absolute scaled error.} MASE is a recently-proposed metric that scales the mean absolute error relative to a naive forecast based on forward propagation of the most recent time series value \cite{hyndman2006another},
\[
\frac{1}{d} \sum_{m=1}^{d} \dfrac{\frac{1}{t} \sum_{t'=1}^{t} |y_{m}(t') - \hat{y}_{m}(t')|}{\frac{1}{t -1} \sum_{t'=2}^{t} |y_{m}(t') - y_{m}(t' - 1)|}
\]
where $m$ indexes the dimensions of a $d$-dimensional trajectory. We find that this metric is prone to divergence at long forecasting times, reducing its interpretability. However, it yields a nearly identical model ranking to the sMAPE metric.

\item \textbf{Coefficient of determination.} The $r^2$ represents the proportion of variance in the original time series that is explained by the predicted time series, averaged across dimensions.
\[
1 - \dfrac{\sum_{m=1}^{d} \sum_{t'=1}^{t} (y_m(t') - \hat{y}_m(t'))^2}    {\sum_{m=1}^{d} \sum_{t'=1}^{t} (y_m(t') - \bar{y}_{m})^2}
\]
where $m$ indexes the dimensions of a $d$-dimensional trajectory and $\bar{y}_m = (1/t) \int_{0}^t y_m(t') dt'$, or equivalently $(1/t) \sum_{t'=1}^t y_m$ for a discrete time series. This quantity has a well-defined upper bound at $1$, and tends to smoothly decay. We find that the general ranking of models remains the same under this method, though it exhibits less dynamic range than sMAPE or Spearman correlation.

\item \textbf{Weighted absolute percent error.} WAPE scales the absolute percent error based on the absolute values of the true time series,
\[
\frac{1}{d} \sum_{m=1}^{d} \frac{\sum_{t'=1}^{t} |y_{m}(t') - \hat{y}_m(t')|}{\sum_{t'=1}^{t} |y_m(t')|}
\]
where $m$ indexes the dimensions of a $d$-dimensional trajectory. This metric has been proposed as a stable metric for comparing forecasts across highly distinct time series, especially those with varying lengths \cite{hewamalage2021look}. In practice, we find that the accrual of errors tends to cause the numerator to diverge at intermediate forecasting horizons. Nonetheless, we find that the ranking of models produced by this metric agrees with other metrics.

\item \textbf{Other metrics: mean squared error (MSE), mean absolute error (MAE), Pearson correlation, Kendall-Tau correlation, mean absolute percentage error (MAPE), Mutual Information (MI), mean absolute ranged relative error (MARRE), root mean squared logarithmic error (RMSLE), and Coefficient of Variation (CV)}. Apart from mutual information, these metrics have very similar properties to those highlighted here, and so we defer discussing them in detail. The mutual information calculation uses recently-introduced density estimation methods \cite{perez2008estimation,kraskov2004estimating,kozachenko1987sample,evans2008computationally,lombardi2016nonparametric}, and it can, in principle, capture nonlinear dependencies between a forecast and the true time series. However, we find it to be too sensitive to fluctuations in time series values to smoothly illustrate forecast quality. 

We note, however, our recent work showing that all of these metrics empirically correlate strongly with sMAPE \cite{gilpin2023recurrences,gilpin2021chaos}, and our forecasting results in terms of these metrics are available in our open-source code repository.

\end{enumerate}

\subsection{Reconstructing invariant measures}

\begin{figure*}
{
\centering
\includegraphics[width=\linewidth]{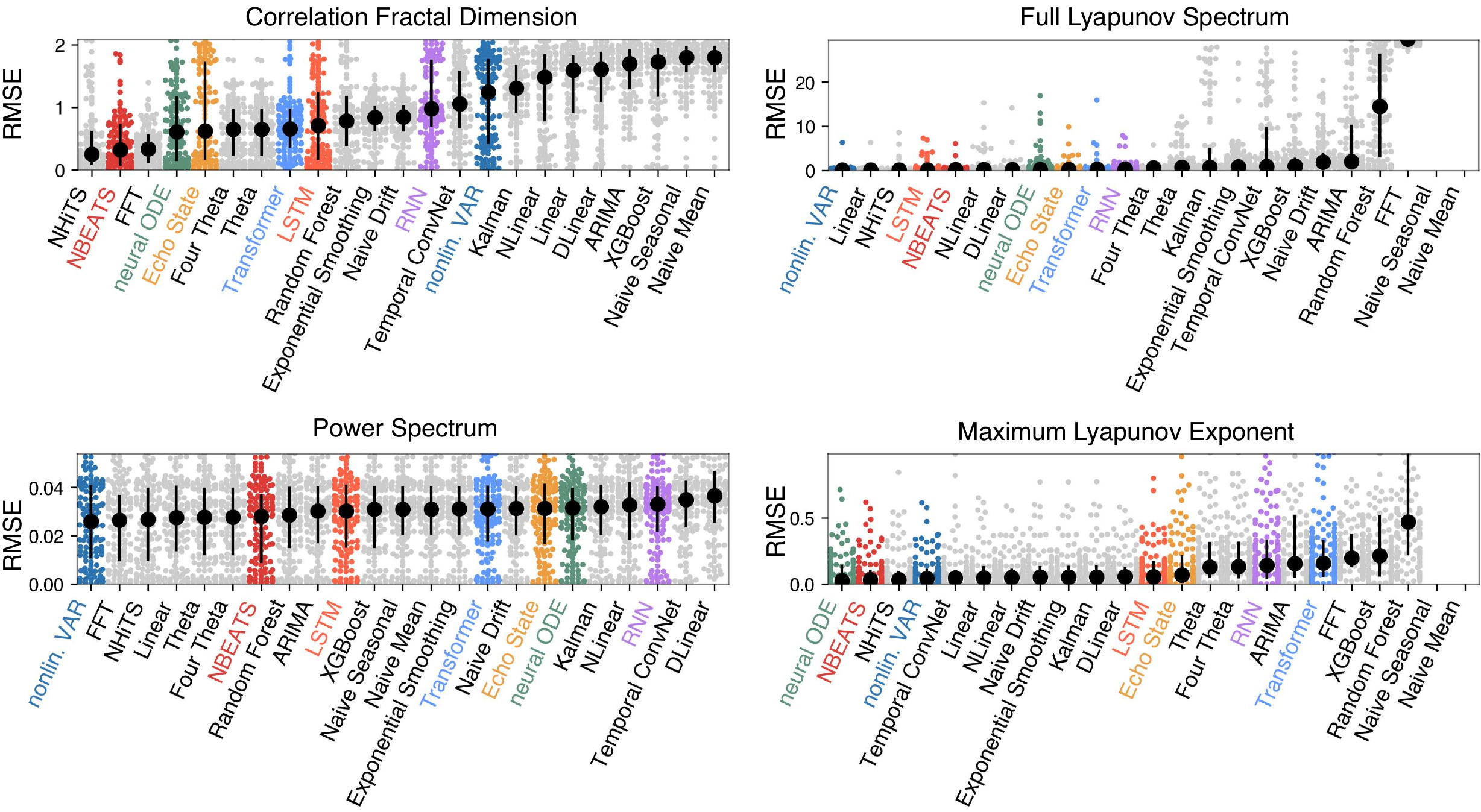}
\caption{
The root-mean-squared error between the true values of various dynamical properties, and the forecasts generated by different forecasting methods. The Lyapunov exponent calculation is ill-defined for constant-valued naive forecasts, leading to missing bars in the Lyapunov exponent spectrum and largest Lyapunov exponent comparisons. Colors correspond to highlighted models from the main text.
}
\label{invariant}
}
\end{figure*}

We further assess quality of forecasts by computing invariant properties of the learned chaotic attractors. In all cases, we compute the invariant property separately on the test dataset's true values and on the forecast generated by each method. For a given invariant measure with ground-truth value $\eta_k$ for the $k^{th}$ dynamical system, we compute an estimate $\hat\eta_{ik}$ for the $i^{th}$ forecasting method using the full predicted time series over the maximum forecast horizon trajectory $t \in (0, T - t^*)$. In Figure \ref{invariant}, we show the error $\abs{\eta_k - \hat{\eta}_{ik}}$ for four different invariant quantities, across all $135$ dynamical systems and $24$ forecast methods.

\begin{enumerate}
\item \textbf{Power spectrum}. Chaotic systems have continuous power spectra due to their fractal nature. By comparing the power spectrum of the original and forecasted systems, we can assess whether forecasts predominantly capture dominant periodic trends in the time series, or whether they capture variation across scales. We find that the nonlinear vector autoregressive model, which is related to echo state networks, most strongly preserves the underlying spectrum. We hypothesize that the fixed nonlinearities in this method allow it to preserve spectral resolution relative to the fully-trainable neural networks, which learn nonlinear relationships from finite data and thus have finite resolution. Other strongly-performing models include the Fourier transform regression model (FFT), which explicitly preserves the power spectrum, as well as simple seasonality models like the Theta family, which model mixed seasonality. Among the general-purpose forecasting models, the related models NBEATS and NHiTS both perform competitively relative to other systems.
\item \textbf{Fractal Dimension}. The fractal dimension quantifies the space-filling properties of an attractor relative to filled solids or planes. We compute the correlation dimension using the robust, non-parametric Grassberger-Procaccia algorithm \cite{grassberger1983characterization}. We find the NBEATS model and its lightweight variant, NHiTS, sharply outperform other methods, suggesting that these methods not only generate pointwise-accurate forecasts but also capture fundamental structural properties of the underlying attractor. We note that the neural ordinary differential equation performs strongly on this metric, despite performing comparatively weakly in absolute pointwise accuracy, suggesting that this model captures the attractor geometry even in the absence of pointwise accuracy.
\item \textbf{Lyapunov Spectrum}. We estimate the full Lyapunov exponent spectrum using standard techniques based on continuous QR factorization of a bundle of tangent vectors transported with the flow \cite{eckmann1986liapunov,abarbanel1991lyapunov}. We note that this algorithm fails when the forecasts are constant (as occurs in the naive models), leading the eigenvalue spectrum of the Jacobian matrix to become singular, and resulting in empty regions on our plot due to the spectrum becoming ill-defined. The nonlinear vector autoregressive model, which is related to reservoir computers, performs strongly on this task, suggesting that the high intrinsic capacity of these models allows them to capture the long-term structure of the underlying attractor. Interestingly, on this particular task, the LSTM model outperforms the NBEATS, in contrast to other tasks.
\item \textbf{Largest Lyapunov exponent}. Rather than consider the full Lyapunov spectrum, we instead compare only the largest Lyapunov exponent $\lambda_\text{max}$ of the reconstructed attractor. This quantity represents a putative measure of chaoticity for each dynamical system. We find that the neural ODE model performs surprisingly strongly on this task, and we speculate that the learning quality of these models (which are trained using adjoint optimization on numerically-integrated trajectories) leads the model to exhibit particular sensitivity to this invariant quantity. Otherwise, for this quantity, as in others, the best-performing models NBEATS and its relative NHiTS represent the strongest baselines.
\end{enumerate}

Our results show that the forecasting methods with the highest pointwise accuracy generally also exhibit the highest accuracy in recovering global invariant properties of the underlying attractors. However, we note that nVAR performed better on this task, which may stem from its inductive bias on the chaotic systems dataset due to the appearance of quadratic kernels in its reservoir \cite{gauthier2021next,zhang2023catch}. Additionally, recent works have shown that both reservoir-based and traditional recurrent models are capable of learning diverse global dynamical properties from high-dimensional chaotic time series when given sufficient training data \cite{margazoglou2023stability,ozalp2023reconstruction}. In particular, these works highlight the ability of certain methods to learn covariant Lyapunov vectors, which encode geometric properties of transport in chaotic flows \cite{margazoglou2023stability,cvitanovic2005chaos}.

\section{Correlation of model performance with invariant properties}
\label{a_embedding}

\begin{figure}
{
\centering
\includegraphics[width=\linewidth]{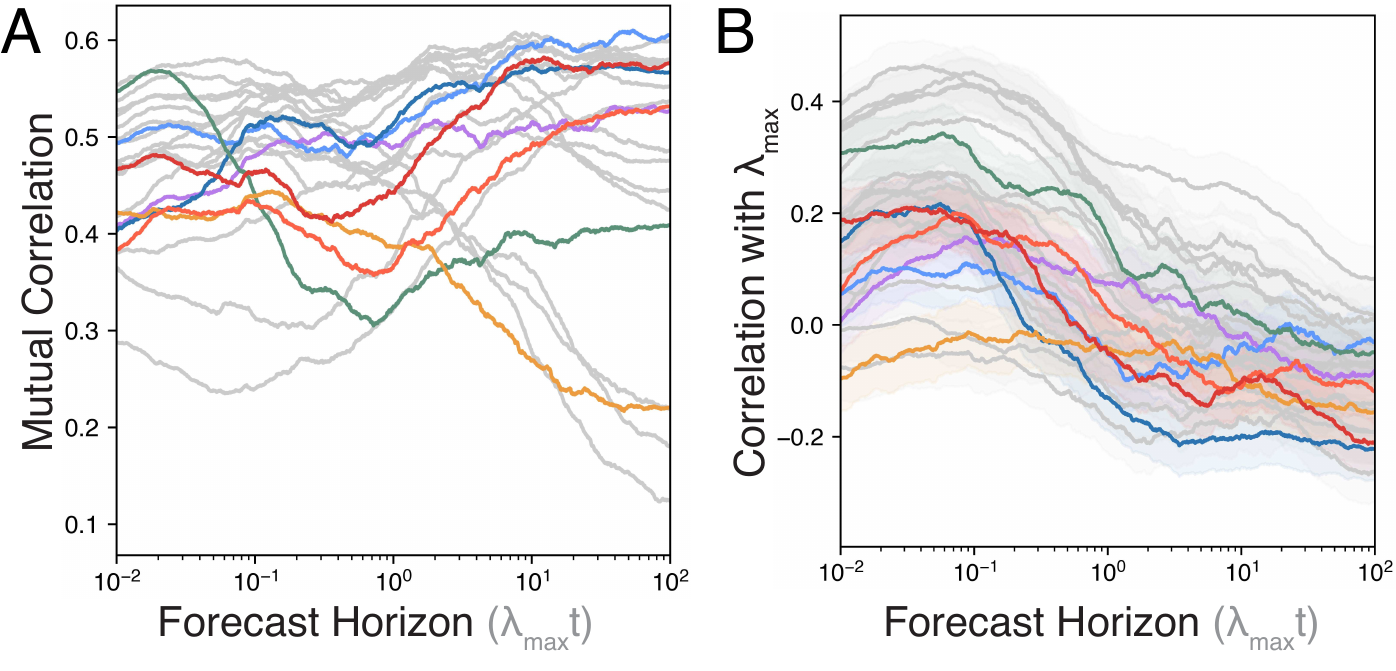}
\caption{
\textbf{Dynamical system properties and forecasting performance.} (A) Mutual correlation of each forecast model with other models, as a function of forecasting horizons. Higher values indicate that the particular model tends to agree with other models regarding which dynamical systems are easier or harder to forecast at that forecasting horizon. (B) Correlation of forecasting performance of each model with the largest Lyapunov exponent of each dynamical system, as a function of forecasting horizon. Colors correspond to highlighted models from the main text, and error bars represent 95\% confidence intervals.
}
\label{correlations}
}
\end{figure}

We first investigate the degree to which each forecasting model tends to agree with other models regarding which dynamical systems are easier or harder to forecast. In Figure \ref{correlations}A we show the mutual correlation $\mathcal{C}_i(t)$ for each model, while in  Figure \ref{correlations}B we correlate model forecasts with the largest Lyapunov exponent $\lambda_\text{max}$ for each dynamical system.

We calculate this quantity as follows: Let $\epsilon_{ik}(t)$ denote the error of the $i^{th}$ forecasting model on the $k^{th}$ dynamical system at future time $t \in [0, T_\text{fut}]$ after the end of training data availability. If there are $K$ total dynamical systems and $N$ forecasting models, then to compute the Spearman correlation we first calculate the ordinal rank variables $R_{ik}(t) \in \{1, 2, ..., N\}$ for each $\epsilon_{ik}(t)$. We may define the time-dependent Spearman correlation matrix $C(t)$ as:
\[
C_{ij}(t) = \frac{\sum_{k=1}^{K} \left[ (R_{ik}(t) - \bar{R}_i(t))(R_{jk}(t) - \bar{R}_j(t)) \right]}{\sqrt{\sum_{k=1}^{K} (R_{ik}(t) - \bar{R}_i(t))^2} \sqrt{\sum_{k=1}^{K} (R_{jk}(t) - \bar{R}_j(t))^2}}
\]
where $\bar{R}_i(t)$ and $\bar{R}_j(t)$ are the time-dependent mean rank variables for $i$ and $j$, respectively.

We define the mutual correlation $\mathcal{C}_i(t)$ for each forecasting model by taking the sum of this quantity over rows,
\[
\mathcal{C}_i(t) = \sum_{j=1}^{K} C_{ij}(t).
\]

In Figure \ref{correlations}A, we find that the high-capacity models that lack structural priors for dynamical systems (NBEATS/NHiTS, transformer, LSTM, RNN) all remain correlated with each other across a wide range of forecasting horizons, and that their degree of mutual correlation increases at long forecasting horizons. In contrast, the echo state networks and neural ODE models, despite being performant overall in the forecasting task, disagree with the remaining models regarding which dynamical systems are easier or harder to forecast, particularly at long forecasting times. 

In order to determine whether this effect can be explained by interactions between forecasting models and invariant properties of different dynamical systems, in Figure \ref{correlations}B, we correlate forecasting results with the largest Lyapunov exponent $\lambda_\text{max}$ for each system. While a weak correlation between forecast error and $\lambda_\text{max}$ holds at short forecasting horizons ($<\lambda_\text{max}^{-1}$), this correlation degrades at long forecasting horizons. This effect implies that the intrinsic chaoticity of different dynamical systems does not determine their empirical predictability over long forecasting horizons, at least when they are learned by modern large forecasting methods.

\section{Timing experiments}

Timing experiments are performed separately for each forecasting method and dynamical system pair, for a total of $135 \times 24 = 3240$ experiments. Our experiment design, implementation, and interpretation closely follows standard methods used in previous works that benchmark time series methods \cite{dempster2020rocket}. Timing experiments are performed on identical hardware restricted to a single CPU core per dataset and per run, with $32$ GB RAM and an AMD EPYC $7763$ Processor ($2.45$ GHz). Timing results are averaged over all $135$ chaotic systems for the performance benchmarks. We note that the presence of GPU and various hardware-level optimizations could improve timing results for larger models with parallelizable training methods, but the underlying number of hardware operations would remain the same.

\section{Embedding the dynamical systems dataset}

We generate $40$ trajectories emanating from distinct random initial conditions on the attractor. Each trajectory has a length $2000$ timepoints, with a sampling rate $100$ points per dominant period as determined by surrogate methods described above. For each system and trajectory, we compute $787$ features based on known signal processing transforms using the \texttt{tsfresh} toolkit \cite{christ2018time}; these include properties such as wavelet coefficients, Friedrich coefficients, and statistical cumulants. The full list of signal features can be found in the \texttt{tsfresh} publication, as well as its accompanying open-source codebase \cite{christ2018time}. We also calculate an  additional $118$ features typically used to characterize the complexity of dynamical time series using the \texttt{neurokit2} toolkit \cite{makowski2021neurokit2,lau2022brain}. These include the various measures of entropy (e.g. sample entropy, permutation entropy), detrended fluctuation analysis, and Hurst exponents. The full list of signal features can be found in the \texttt{neurokit2} publication, as well as its accompanying open-source codebase \cite{makowski2021neurokit2,lau2022brain}.  Across all dynamical systems and trajectories, we subselect only the features with greater variance across different dynamical systems than across replicate trajectories within each system, resulting in a vector containing $747$ informative features representing each dynamical system. 

We embed each trajectory by using the uniform manifold approximation and projection (UMAP) nonlinear embedding technique, which seeks to represent the original, high-dimensional feature vectors in a lower-dimensional space that preserves local topology and nearest neighbors \cite{mcinnes2018umap}. For each dynamical system, we compute the median position in the reduced-order UMAP space as the exemplary position of that particular system.

\section{Invariant property calculation}

For each system in the dataset, the largest Lyapunov exponent $\lambda_\text{max}$ is calculated using multiple methods in order to ensure accuracy. The first method continuously calculates the Jacobian of the dynamical equation along a trajectory using its analytical expression, which can be quickly calculated using modern automatic differentiation software \cite{baydin2018automatic}. Given a time-dependent Jacobian matrix along a trajectory, the full Lyapunov spectrum can be calculated using composition of the instantaneous QR factorization of the matrix at each timepoint \cite{eckmann1986liapunov,abarbanel1991lyapunov}. As a secondary check, we also calculate the largest Lyapunov exponent using a naive method based purely on the classical definition of the Lyapunov exponent, $\lambda_{max} = \lim_{t\rightarrow\infty} \log(||\v{x}(t) - \v{x'}(t) ||_2 / ||\v{x}(0) - \v{x'}(0)||_2)$, where $\v{x}(0) = \v{x}_0$ and $\v{x'}(0) = \v{x}_0 (1 + \gv{\xi})$. In practice, we set $||\gv{\xi}||_2 \leq 10^{-14}$, a quantity well above the floating-point precision floor, and stop the calculation when $||\v{x}(t) - \v{x'}(t) ||_2 > 10^{-8}$.

For both the QR factorization and naive methods, we perform two calculations for consistency: a long-time calculation in which we average the Lyapunov exponent estimates along $1000$ distinct trajectories each of length $5000$ timepoints, and a short-time calculation in which we average Lyapunov exponent estimates along $50000$ distinct trajectories each of length $100$. The two calculations agree to within two significant figures, validating that integration steps and durations are sufficient to reach the ergodic limit of each dynamical system.

Having obtained the largest Lyapunov exponents, we estimate other invariant properties using the same procedures. The correlation fractal dimension is estimated for each attractor using the Grassberger-Procaccia algorithm \cite{grassberger1983characterization}, and the multiscale entropy is estimated using algorithms for multivariate time series \cite{ahmed2011multivariate}. The Kaplan-Yorke fractal dimension is estimated directly from the Lyapunov exponent spectrum.

%\clearpage
\bibliography{forecast_cites} 
\bibliographystyle{naturemag}

\end{document}